\documentclass[conference,a4paper]{IEEEtran}
\IEEEoverridecommandlockouts

\usepackage{fancyhdr}
\usepackage[dvips]{graphicx}
\usepackage{indent}
\usepackage{amsmath,amssymb}
\usepackage{paralist}
\usepackage{eclbkbox}
\usepackage{subfigure}
\usepackage{multirow}

\begin{document}
\title{An Estimation of Favorite Value in Emotion Generating Calculation by Fuzzy Petri Net
\thanks{\copyright 2013 IEEE. Personal use of this material is permitted. Permission from IEEE must be obtained for all other uses, in any current or future media, including reprinting/republishing this material for advertising or promotional purposes, creating new collective works, for resale or redistribution to servers or lists, or reuse of any copyrighted component of this work in other works.}
}

\author{\IEEEauthorblockN{Takumi Ichimura}
\IEEEauthorblockA{Faculty of Management and Information Systems,\\
Prefectural University of Hiroshima\\
1-1-71, Ujina-Higashi, Minami-ku,\\
Hiroshima, 734-8559, Japan\\
Email: ichimura@pu-hiroshima.ac.jp}
\and
\IEEEauthorblockN{Kousuke Tanabe}
\IEEEauthorblockA{Graduate School of Comprehensive Scientific Research,\\
Prefectural University of Hiroshima\\
1-1-71, Ujina-Higashi, Minami-ku,\\
Hiroshima, 734-8559, Japan\\
Email: bakabonn009@gmail.com}
}

\maketitle

\fancypagestyle{plain}{
\fancyhf{}	
\fancyfoot[L]{}
\fancyfoot[C]{}
\fancyfoot[R]{}
\renewcommand{\headrulewidth}{0pt}
\renewcommand{\footrulewidth}{0pt}
}

\pagestyle{fancy}{
\fancyhf{}
\fancyfoot[R]{}}
\renewcommand{\headrulewidth}{0pt}
\renewcommand{\footrulewidth}{0pt}

\begin{abstract}
Emotion Generating Calculations (EGC) method based on the Emotion Eliciting Condition Theory can decide whether an event arouses pleasure or not and quantify the degree under the event. An event in the form of Case Frame representation is classified into 12 types of calculations. However, the weak point in EGC is Favorite Value ($FV$) as the personal taste information. In order to improve the problem, this paper challenges to establish a learning method to learn speaker's taste information from dialog. Especially, the learning method employs Fuzzy Petri Net to find an appropriate $FV$ to a word which has the unknown $FV$. This paper discusses the effective learning method to improve a weak point of EGC when a missing value of $FV$ exists.
\end{abstract}


\IEEEpeerreviewmaketitle

\section{Introduction}
\label{sec:Introduction}
Our research group proposed an estimation method to calculate the agent's emotion from the contents of utterances and to express emotions which are aroused in computer agent by using synthesized facial expression \cite{Ichimura03, Mera02, Mera03}. Emotion Generating Calculations (EGC) method \cite{Mera03} based on the Emotion Eliciting Condition Theory \cite{Elliott92} can decide whether an event arouses pleasure or not and quantify the degree of pleasure under the event.

Calculated emotions by EGC will change the mood of the agent. Ren \cite{Ren06} describes Mental State Transition Network (MSTN) which is the basic concept of approximating to human psychological and mental responses. The assumption of discrete emotion state is that human emotion is classified into some kinds of stable discrete states, called ``mental state,'' and the variance of emotions occurs in the transition from a state to other state with an arbitrary probability. Mera and Ichimura \cite{Mera10, Ichimura13} developed a computer agent that can transit a mental state in MSTN based on analysis of emotion by EGC method. EGC calculates emotion and the type of the aroused emotion is used to transit mental state \cite{Mera10}.

However, the weak point in EGC is Favorite Value ($FV$) as the personal taste information. In order to improve the problem, Mera et al. challenged to establish a method to learn speaker's taste information from dialog \cite{Mera01}. The learning method consists of four following parts:(1)Directly expression according to the degree of like/dislike, (2)Decision Making based on the speaker's environment, (3)Relevance to the degree of displeasure to the object, and (4)Backward calculation from the emotional expression. The method seems to be effective in theory, but the procedure (2) is still an theoretical query into emotions, because it does not include the reasoning model. In this paper, the learning method employs Fuzzy Petri Net (FPN) to find an appropriate $FV$ for a word which has the unknown $FV$. The $FV$ given by the method can change the value itself according to the meaning or intent of discourse in a dialog. Moreover, the reasoning method can control the degree of changing $FV$, because the threshold value in FPN varies as the mood in MSTN changes. This paper discusses the effective learning method to improve a weak point of EGC when a missing value of $FV$ exists.

The remainder of this paper is organized as follows. In the section \ref{sec:EGC}, the brief explanation to understand the EGC is described. Section \ref{sec:MentalStateTransitionLarningNetwork} explains to MSTN to measure the current mental state by the stimulus of the EGC. Section \ref{sec:FuzzyPetriNet} describes the Fuzzy Petri Net model and Section \ref{sec:LearningFV} proposes the reasoning method and learning method of $FV$ modification. In Section \ref{sec:ConclusiveDiscussion}, we give some discussions to conclude this paper.

\section{Emotion Generating Calculations}
\label{sec:EGC}
\subsection{An Overview of Emotion Generating Process}

\begin{figure}[ht]
\begin{center}
\includegraphics[scale=0.5]{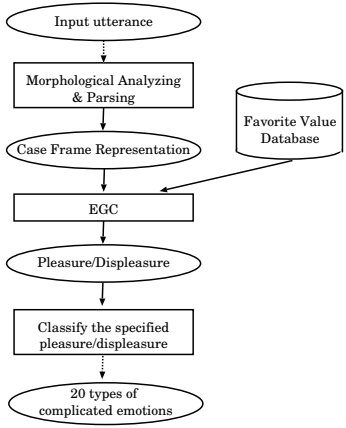}
\caption{Process for generating emotions}
\label{fig:processgeneratingemotion}
\vspace{-0.3cm}
\end{center}
\end{figure}

Fig.\ref{fig:processgeneratingemotion} shows the emotion generating process where the user's utterance is transcribed into a case frame representation based on the results of morphological analysis and parsing. The agent works to determine the degree of pleasure/displeasure from the event in case frame representation by using EGC. EGC consists of 2 or 3 terms such as  subject, object and predicate, which have Favorite Value ($FV$), the strength of the feelings described in section \ref{sec:FavoriteValue}.

Then, the agent divides this simple emotion (pleasure/displeasure) into 20 various emotions based on the Elliott's ``Emotion Eliciting Condition Theory\cite{Elliott92}.'' Elliott's theory requires judging conditions such as ``feeling for another,'' ``prospect and confirmation,'' and ``approval/disapproval.'' The detail of this classification method is described in the section \ref{sec:ComplicatedEmotion}.

\subsection{Favorite Value Database}
\label{sec:FavoriteValue}
Which an event is pleasure or displeasure is determined by using $FV$. $FV$ is a positive/negative number to an object when the user likes/dislikes it, respectively.$FV$ is predefined a real number in the range $[-1.0, 1.0]$. There are two types of $FV$s, personal $FV$ and initial $FV$. Personal $FV$ is stored in a personal database for each person who the agent knows well, and it shows the degree of like/dislike to an object from the person's viewpoint. On the other hand, initial $FV$ shows the common degree of like/dislike to an object that the agent feels. Generally, it is generated based on the agent's own preference information according to the result of some questionnaires. Both personal and initial $FV$s are stored in the user own database. An initial value of $FV$ is determined beforehand on the basis of `corpus' of its applied field. The $FV$s of the objects are gained from a questionnaire. However, there are countless objects in the world. In this paper, we limit the objects that have initial $FV$ into the frequently appeared words in dialog.

\subsection{Equation of EGC}
We assume an emotional space as three-dimensional space. Therefore, we present a method to distinguish pleasure/displeasure from an event by judging the existence of `synthetic vector''\cite{Mera02}.

\begin{table}[tbp]
\begin{center}
\caption{Correspondence between the event type and the axis}
\begin{tabular}{c|c|c|c}
\hline
Event type  & $f_{1}$ & $f_{2}$ & $f_{3}$ \\ \hline 
$V(S)$      &         &        &         \\
$A(S,C)$    &         &        &         \\
$A(S,OF,C)$ &         &        &         \\
$A(S,OT,C)$ & $f_{S}$  &        & $f_{P}$ \\
$A(S,OM,C)$ &         &        &         \\
$A(S,OS,C)$ &         &        &         \\ \hline
$V(S,OF)$   & $f_{S}$ & $f_{OT}-f_{OF}$ & $f_{P}$  \\
$V(S,OT)$   &         &         &         \\ \hline
$V(S,OM)$   & $f_{S}$ & $f_{OM}$ & $f_{P}$  \\ \hline
$V(S,OS)$   & $f_{S}-f_{OS}$ &  & $f_{P}$  \\ \hline
$V(S,O)$    & $f_{S}$ & $f_{O}$ & $f_{P}$  \\ 
            & $f_{O}$ &         & $f_{P}$ \\ \hline
$V(S,O,OF)$ & $f_{O}$ & $f_{OT}-f_{OF}$ & $f_{P}$ \\
$V(S,O,OT)$ &         & $f_{OM}$       &  \\ \hline
$V(S,O,OM)$ & $f_{O}$ & $f_{OM}$ & $f_{P}$ \\ \hline
$V(S,O,I)$ & $f_{O}$ & $\mid f_{I} \mid$  & $f_{P}$ \\ \hline
$V(S,O,OC)$ & $f_{O}$ &          & $f_{OC}$ \\ \hline
$A(S,O,C)$  & $f_{O}$ &          & $f_{P}$ \\ \hline
\end{tabular}
\label{tab:EGC-eventtype}
\vspace{-0.3cm}
\end{center}
\end{table}

Table \ref{tab:EGC-eventtype} shows the correspondence between the case element in EGC equations and the axis in the three-dimensional model. In Table \ref{tab:EGC-eventtype}, `V(S,*)' is the type of event (verb) and `A(S,*)' is the type of attribute (adjective). the variables denoted in Table \ref{tab:EGC-eventtype} are expressed as follows.

\begin{itemize}
\item $f_{S}$ : $FV$ of Subject
\item $f_{OF}$ : $FV$ of Object-From
\item $f_{OM}$ : $FV$ of Object-Mutual
\item $f_{OC}$ : $FV$ of Object-Content
\item $f_{O}$ : $FV$ of Object
\item $f_{OT}$ : $FV$ of Object-To
\item $f_{OS}$ : $FV$ of Object-Source
\item $f_{P}$ : $FV$ of Predicate
\item $f_{I}$ : $FV$ of Instrument or tool
\end{itemize}

\begin{figure}[btp]
\begin{center}
\includegraphics[scale=0.35]{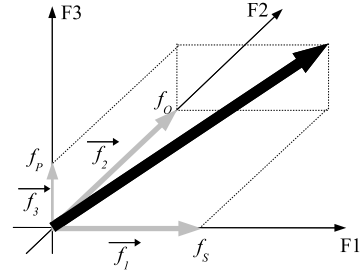}
\caption{Emotion Space for EGC}
\label{fig:EGC-emotionvector}
\vspace{-0.2cm}
\end{center}
\end{figure}

Table \ref{tab:EGC-axis} shows the relation between the sign of axis in each dimension and the pleasure/displeasure of generated emotion. When the vector is on the axis, the event does not raise any emotion. When we calculate the synthetic vectors of the events which do not have $f_{i}$ elements, we supply a dummy $FV$, $\beta$ as $f_{i}$ element. We tentatively defined $\beta$ as $+0.5$. Fig.\ref{fig:EGC-emotionvector} is an example of emotion space of event type $V(S, O)$. There are three elements, Subject, Object, and Predicate, in the event type, and the orthogonal vectors by the elements construct a rectangular solid. 

\begin{table}[tbp]
\begin{center}
\caption{pleasure/displeasure in emotional space}
\begin{tabular}{c|c|c|c|c}
\hline
Area & $f_{1}$ & $f_{2}$ & $f_{3}$ & Emotion \\ \hline
I & + & + & + & Pleasure \\
I\hspace{-.1em}I & - & + & + & Displeasure \\
I\hspace{-.1em}I\hspace{-.1em}I & - & - & + & Pleasure \\
I\hspace{-.1em}V & + & - & + & Displeasure \\
V & + & + & - & Displeasure \\
V\hspace{-.1em}I & - & + & - & Pleasure \\
V\hspace{-.1em}I\hspace{-.1em}I & - & - & - & Displeasure \\
V\hspace{-.1em}I\hspace{-.1em}I\hspace{-.1em}I & + & - & - & Pleasure \\ \hline
\end{tabular}
\label{tab:EGC-axis}
\vspace{-0.3cm}
\end{center}
\end{table}

\subsection{Complicated Emotion Eliciting Method}
\label{sec:ComplicatedEmotion}
\begin{figure}[btp]
\begin{center}
\includegraphics[scale=0.7]{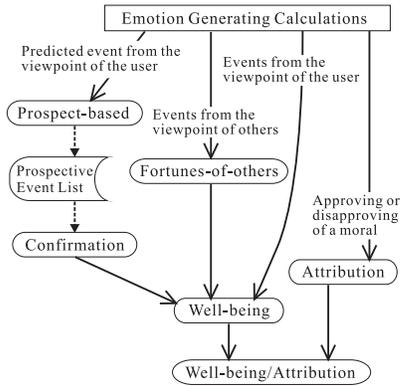}
\caption{Dependency among emotion groups}
\label{fig:EGC}
\vspace{-0.3cm}
\end{center}
\end{figure}

Based on emotion values calculated by EGC method and their situations, the pleasure/displeasure is classified into 20 types of emotion. We consider only 20 emotion types, which are classified into six emotional groups as follows, ``joy'' and ``distress'' as a group of ``Well-Being,'' ``happy-for,'' ``gloating,'' ``resentment,'' and ``sorry-for'' as a group of ``Fortunes-of-Others,'' ``hope'' and ``fear'' as a group of ``Prospect-based,'' ``satisfaction,'' ``relief,'' ``fears-confirmed,'' and ``disappointment'' as a group of ``Confirmation,'' ``pride,'' ``admiration,'' ``shame,'' and ``disliking'' as a group of ``Attribution,'' ``gratitude,'' ``anger,'' ``gratification,'' and ``remorse'' as a group of ``Well-Being/Attribution'' \cite{Mera03}. Fig.\ref{fig:EGC} shows the dependency among the groups of emotion types. 

\section{Mental State Transition Learning Network}
\label{sec:MentalStateTransitionLarningNetwork}
\subsection{Mental State Transition Network} 
MSTN, proposed by Ren\cite{Ren06}, represents the basic concept of approximating to human physiological and mental responses. The method focuses not only information included in the elements of phonation, facial expressions, and speech, but also human psychological characteristics based on the latest achievements of brain science and psychology in order to derive transition networks for human psychological states. The assumption of discrete emotion state is that human emotions are classified into some kinds of stable discrete states, called ``mental state'', and the variance of emotions occurs in the transition from a state to other state with a probability. The probability of transition is called ``transition cost'' and it does not have the same one. Moreover, with no stimulus from the external world, the probability may converge to fall into a certain value as if the confusion of the mind leaves and is relieved. On the contrary, with a stimulus from external world and/or attractive thought in internal world, the continuous accumulated emotional energy cannot jump to the next mental state and remains in its mental state still. The simulated model of mental state transition network\cite{Ren06} describes the simple relations among some kinds of stable emotions and the corresponding transition probability. The probability was calculated from analysis of many statistical questionnaire data.

\begin{figure}[tbp]
\begin{center}
\subfigure[An Overview of MSTN]{
\includegraphics[scale=0.3]{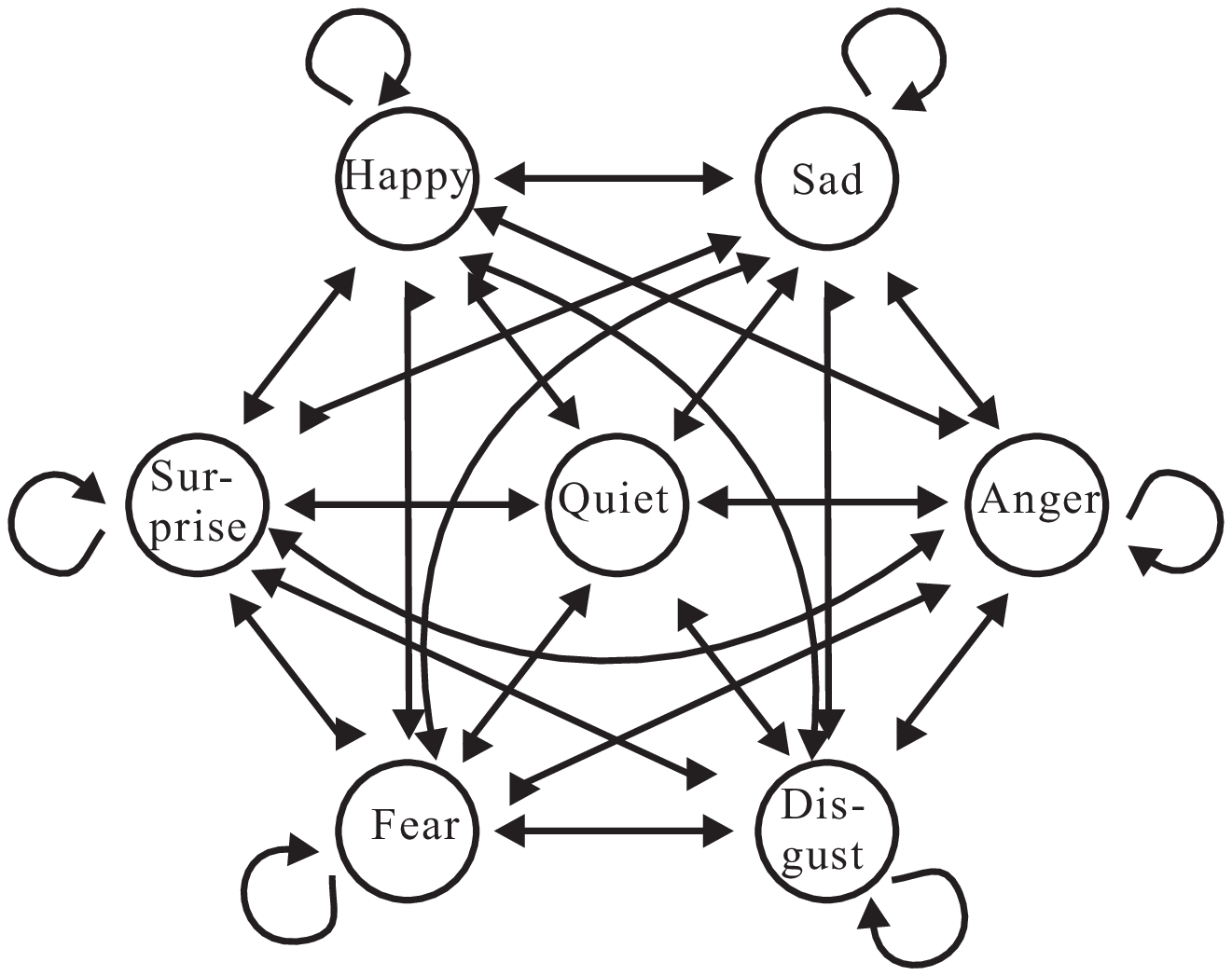}
\label{fig:MSTN}
}
\subfigure[Transition Cost]{
\includegraphics[scale=0.3]{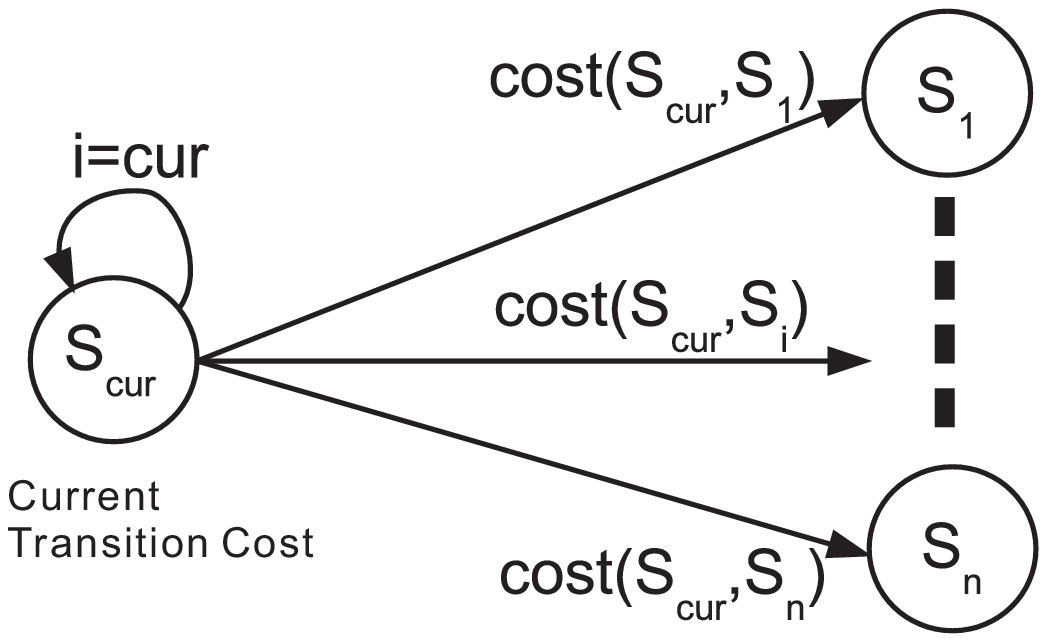}
\label{fig:TransitionCost}
}
\caption{MSTN model}
\label{fig:MSTN model}
\end{center}
\end{figure}

As shown in Fig.\ref{fig:MSTN}, the MSTN denotes a mental state as a node, a set of some kinds of mental state $\mathcal{S}$, the current emotional state $\mathcal{S}_{cur}$ , and the transition cost $cost(\mathcal{S}_{cur}, \mathcal{S}_{i})$ as shown in Fig.\ref{fig:TransitionCost}.

In \cite{Ren06}, six kinds of mental states and quiet state are considered for questionnaire. That is, the transition table of $cost(\mathcal{S}_{i}, \mathcal{S}_{j})$, $i=1,2,\cdots,7$, $j=1,2,\cdots,7$ is prepared. The experiment for participants was examined without stimulus from external world. Each participant fills in the numerical value from 1 to 10 that means the strength of relation among mental states. Moreover, the same questionnaire was examined under the condition with the stimulus from external world. The 200 participants answered the questionnaire. The numerical values in Table \ref{tab:TransitionCost} show the statistical analysis results. The symbols in Table \ref{tab:TransitionCost} mean $s_{1}$=`happy', $s_{2}$=`quiet', $s_{3}$=`sad', $s_{4}$=`surprise', $s_{5}$=`angry', $s_{6}$=`fear', $s_{7}$=`disgust', respectively. The transition cost from each current state to the next state is summarized to $1.0$.

\begin{table}[tbp]
\begin{center}
\caption{Transition Probability in MSTN}
\begin{tabular}{l|c|ccccccc}
\hline
\multicolumn{2}{c|}{ }      & \multicolumn{7}{c}{next}\\
\cline{3-9}
\multicolumn{2}{c|}{ }      & $s_{1}$ & $s_{2}$ & $s_{3}$   & $s_{4}$ & $s_{5}$ & $s_{6}$ & $s_{7}$\\ \hline
\multirow{7}{*}{\rotatebox{90}{current}} &$s_{1}$   & 0.421 & 0.362 & 0.061 & 0.060 & 0.027 & 0.034 & 0.032\\
&$s_{2}$ & 0.213 & 0.509 & 0.090 & 0.055 & 0.039 & 0.051 & 0.042\\
&$s_{3}$ & 0.084 & 0.296 & 0.320 & 0.058 & 0.108 & 0.064 & 0.068\\
&$s_{4}$ & 0.190 & 0.264 & 0.091 & 0.243 & 0.086 & 0.076 & 0.048\\
&$s_{5}$ & 0.056 & 0.262 & 0.123 & 0.075 & 0.293 & 0.069 & 0.121\\
&$s_{6}$ & 0.050 & 0.244 & 0.137 & 0.101 & 0.096 & 0.279 & 0.092\\
&$s_{7}$ & 0.047 & 0.252 & 0.092 & 0.056 & 0.164 & 0.075 & 0.313\\ \hline
\end{tabular}
\label{tab:TransitionCost}
\vspace{-0.3cm}
\end{center}
\end{table}

\subsection{EGC in MSTN}
Even if there are not any signal from external world, the mental state will not change. In this case, the transition probabilities represented in Table \ref{tab:TransitionCost} are adopted to calculate by using EGC.
In this paper, we assume that the stimulus from external world is the utterance of the user and the transition cost is calculated as follows.
\begin{equation}
cost(\mathcal{S}_{i}, \mathcal{S}_{j})=1-\frac{\#(\mathcal{S}_{i} \rightarrow \mathcal{S}_{j})}{\sum_{j=1}^{7} \#(\mathcal{S}_{i} \rightarrow \mathcal{S}_{j})},
\label{eq:TransitionCost}
\end{equation}
where $\#(\mathcal{S}_{i} \rightarrow \mathcal{S}_{j})$ is the number of transition from mental state $\mathcal{S}_{i}$, $1 \leq i \leq 7$ to $\mathcal{S}_{j}$, $1 \leq j \leq 7$. The transition cost is calculated by using the total of $\#(\mathcal{S}_{i} \rightarrow \mathcal{S}_{j})$ for all mental state. Eq.(\ref{eq:TransitionCost}) means that the higher transition cost is, the less transition occurs.

\begin{figure}[btp]
\begin{center}
\includegraphics[scale=0.3]{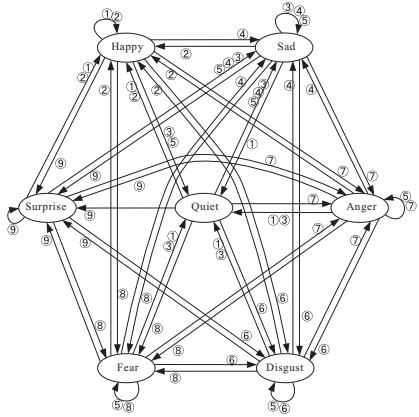}
\caption{MSTN with EGC}
\vspace{-0.8cm}
\label{fig:MSTN_EGC}
\end{center}
\end{figure}

Eq.(\ref{eq:TransitionCost_next}) calculates the next mental state from the current mental state $\mathcal{S}_{cur} \in \mathbf{S}$ by using the emotion vector.
\begin{equation}
next=\arg \max_{k} \frac{e_{k}}{cost(\mathcal{S}_{cur}, \mathcal{S}_{i})}, \ 1 \leq k \leq 9
\label{eq:TransitionCost_next}
\end{equation}
The emotion vector consists of 9 kinds of emotion groups which are classified 28 kinds of emotions as shown in Table \ref{tab:classofemotion}. Fig.\ref{fig:MSTN_EGC} shows the MSTN by using EGC. The circled numbers in Fig.\ref{fig:MSTN_EGC} are the number in the left side of Table \ref{tab:classofemotion}. The $e_{k}$ $(1 \leq k \leq 9)$ shows the strength of emotion group $k$ and takes the maximum value of elements belonged in each set $e_{k}$ as follows.

\begin{description}
\item $e_{1}=\max (e_{gloating}, e_{hope}, \cdots, e_{shy})$
\item $e_{2}=\max (e_{joy}, e_{happy\_for})$
\item $\vdots$
\item $e_{9}=\max (e_{surprise})$

\end{description}
\begin{table}[tbp]
\begin{center}
\caption{Classification of Generated Emotion}
\begin{tabular}{c|c}
\hline
No. & Emotion \\ \hline
  & gloating, hope, satisfaction, relief, pride,\\
1 & admiration, liking, gratitude, gratification,love, shy\\ \hline
2 & joy,  happy\_for \\ \hline
3 & sorry-for, shame, remorse \\ \hline
4 & fear-confirmed, disappointment, sadness \\ \hline
5 & distress, perplexity \\ \hline
6 & disliking, hate \\ \hline
7 & resentment, reproach, anger \\ \hline
8 & fear \\ \hline
9 & surprise \\ \hline
\end{tabular}
\label{tab:classofemotion}
\end{center}
\end{table}

The $emo$ in Eq.(\ref{eq:selectemotion}) calculates the maximum emotion group according to the transition cost between current state and next state.
\begin{equation}
emo_{k}=\arg \max_{k} \frac{e_{k}}{cost(\mathcal{S}_{cur}, next(\mathcal{S}_{cur},k))},  \ 1 \leq k \leq 9, 
\label{eq:selectemotion}
\end{equation}
where $next(\mathcal{S}_{cur}, k)$ is next mental state from the current state by selecting emotion group $k$.

\section{Fuzzy Petri Net}
\label{sec:FuzzyPetriNet}
The interactive system for tourist concierge is implemented by the goal driven reasoning. The inference technique which uses IF-Then rules to repetitively breaks a goal into smaller sub-goals. It is an efficient way to solve problems that can be modeled as ``structured selection problems.'' The aim of the system is to pick the choice from many enumerated possibilities. 

The goal driven reasoning can deduce the user's requirement from the conversation in the specified goal driven reasoning such as tourist information system. Then, IF-THEN rules for the interactive system as knowledge representations are prepared according to the sightseeing spots and goods.

A fuzzy production rule is a rule which describes the fuzzy relation between 2 proposition. Let $R$ b a set of fuzzy production rule $R=\{R_{1}, R_{2}, \cdots, R_{n}\}$. The general form of the $i$th fuzzy production rule $R_{i}$ is as follows:
\begin{equation}
R_{i}:IF\; d_{j}\; Then\; d_{k},\; CF=\mu_{j}
\label{eq:Rule-0}
\end{equation}

 A Fuzzy Petri Net (FPN)\cite{Chen91} is an effective model to implement goal driven reasoning. A FPN structure is defined as 8-tuple\cite{Chen91};
\begin{equation}
FPN = \{P, T, D, I, O, f, \alpha, \beta\},
\label{eq:FPN0-1}
\end{equation}
where $P=\{p_{1}, p_{2}, \cdots, p_{n}\}$ is a finite set of places, $T=\{t_{1}, t_{2}, \cdots, t_{m} \}$ is a finite set of transitions, $D=\{d_{1}, d_{2}, \cdots, d_{n} \}$ is a finite set of propositions, $P \cap T \cap D = \phi$ and $|P|=|D|$.  $I:T\rightarrow P^{\infty}$ is the input function, a mapping from transition to bags of places. $O: T\rightarrow P^{\infty}$ is the output function, a mapping from transition to bags of places. $f: T\rightarrow [0,1]$ is an association function, a mapping from transitions to real values in [0, 1]. $\alpha: P\rightarrow [0,1]$ is an association function, a mapping from places to real values in [0,1]. $\beta: P\rightarrow D$ is an association function, a bijective mapping from laces to propositions.

 Let $A$ be a set of directed arcs. If $p_{j} \in I(t_{i})$, then there exists a directed arc $a_{ji} (\in A)$ from the place $p_{j}$ to the transition $t_{j}$. if $p_{k} \in O(t_{i})$, then there exists a directed arc $a_{ik} (\in A)$ from the transition $t_{i}$ to place $p_{k}$. If $f(t_{i})=\mu_{i}$, $(\mu_{i} \in [0,1])$, then the transition $t_{i}$ is said to be associated with a real value $\mu_{i}$. If $\beta(p_{i})=d_{i}, (d_{i} \in D)$, then the place $p_{i}$ is said to be associated with the proposition $d_{j}$.

The token value in a place $p_{i} (\in P)$ is denoted by $\alpha(p_{i})$, where $\alpha(p_{i}) \in [0.1]$. If $\alpha(p_{i}) = y_{i} (y_{i} \in [0,1])$ and $\beta(p_{i})=d(_{i})$, then it indicates that the degree of truth of proposition $d_{i}$ is $y_{i}$.

\begin{figure}[!tb]
\begin{center}
\includegraphics[scale=0.5]{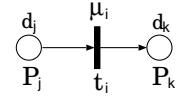}
\caption{Fuzzy Petri Net of If-Then rule}
\label{fig:FPN_simple}
\end{center}
\end{figure}

By using a FPN, the fuzzy production rule such as Eq.(\ref{eq:Rule-0}) can be modeled as shown in Fig.\ref{fig:FPN_simple}. The FPN model has places, transitions, and tokens. A transition may be enabled to fire. A transition $t_{i}$ is enabled if for all $p_{j} (\in I(t_{i})), \alpha(p_{j})\geq \lambda$, where $\lambda$ is a threshold value in $[0,1]$. A transition $t_{i}$ fires by removing the tokens from its input places and then depositing one token into each of its output places. The token value in an output place of $t_{i}$ is calculated by using  Eq.(\ref{eq:token}). \begin{equation}
y_{k}=y_{i} \cdot \mu_{i}
\label{eq:token}
\end{equation}
If there are 2 or more fuzzy variables in the antecedent part of rules, the production of $\min$ of them and the transition by the fuzzy reasoning as follows.
\begin{equation}
y_{k}=\min (y_{j1}, \cdots, y_{jn}) \cdot \mu_{ij}
\label{eq:Min}
\end{equation}

\begin{figure}[tbp]
\begin{center}
\subfigure[Type1]{
\includegraphics[scale=0.57]{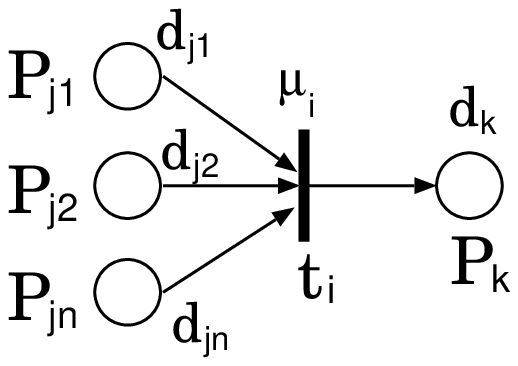}
\label{fig:FPN_1}
}
\subfigure[Type2]{
\includegraphics[scale=0.57]{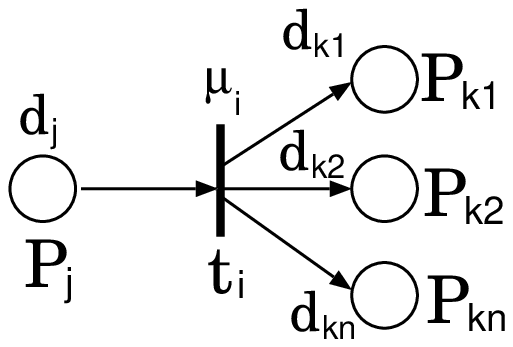}
\label{fig:FPN_2}
}
\subfigure[Type3]{
\includegraphics[scale=0.57]{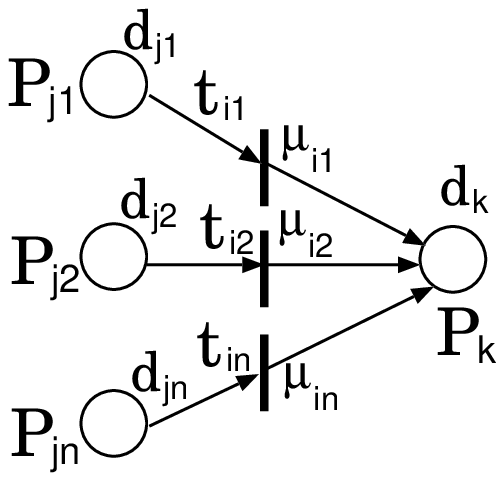}
\label{fig:FPN_3}
}
\subfigure[Type4]{
\includegraphics[scale=0.57]{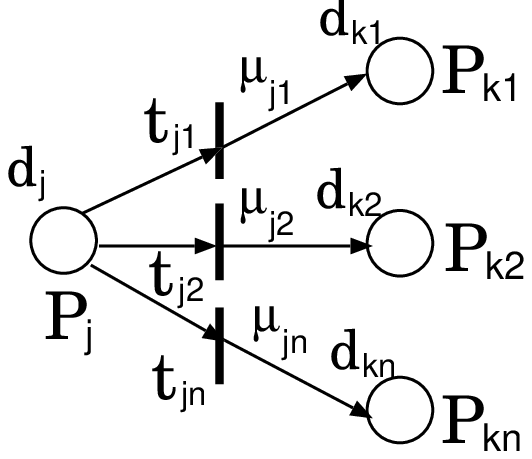}
\label{fig:FPN_4}
}
\caption{FPN models}
\vspace{-0.5cm}
\label{fig:FPN_models}
\end{center}
\end{figure}

Furthermore, the following 4 types of IF-THEN rules by extending Eq.(\ref{eq:Rule-0}) are defined in Eq.(\ref{eq:FPN_rules}). Fig.\ref{fig:FPN_models} shows the FPN model of Eq.(\ref{eq:FPN_rules}). However, Type4 cannot derive clear implication, and then we don’t consider in this paper. By Eq.(\ref{eq:Min2}), the token values are calculated respectively.

\begin{eqnarray}
\nonumber TYPE 1:&&\; {\rm IF}\; d_{j1}\; {\rm and}\; d_{j2}\; \cdots\; {\rm and}\; d_{jn}\; {\rm Then}\; d_{k}, CF=\mu_{i}\\
\nonumber TYPE 2:&&\; {\rm IF}\; d_{j}\; {\rm Then}\; d_{k1}\; {\rm and}\; d_{k2}\; \cdots\; {\rm and}\; d_{kn},\! CF = \mu_{i} \\
\nonumber TYPE 3:&&\; {\rm IF}\; d_{j1}\; {\rm or}\; d_{j2}\; \cdots\; {\rm or}\; d_{jn}\; {\rm Then}\; d_{k},\\
\nonumber \; &&\!CF = \{\mu_{i1}, \mu_{i2}, \cdots, \mu_{in}\} \\
\nonumber TYPE 4:&&\; {\rm IF}\; d_{j}\;  {\rm Then}\; d_{k1}\; \cdots\; {\rm or}\; d_{k2}\; \cdots\; {\rm or}\; d_{kn},\\
\; &&\!CF = \{\mu_{j1}, \mu_{j2}, \cdots, \mu_{jn}\}
\label{eq:FPN_rules}
\end{eqnarray}
\begin{eqnarray}
\nonumber y_{k} &=& \min( y_{j1}, y_{j2}, \cdots, y_{jn} ) \cdot \mu_{i}\\
\nonumber y_{kl} &=& y_{j} \cdot \mu_{i}, \: (1 \leq l \leq n )\\
y_{k} &=& \max( y_{j1} \cdot \mu_{i1}, y_{j2} \cdot \mu_{i2}, \cdots, y_{jn} \cdot \mu_{in})
\label{eq:Min2}
\end{eqnarray}

\section{A learning of Personal Favorite Value}
\label{sec:LearningFV}
\subsection{4 learning procedures of FV}
Although EGC is an effective method to estimate the user's emotion, the weak point in EGC is Favorite Value ($FV$) itself as the personal taste information. In order to improve the problem, Mera et al. challenged to establish a method to learn speaker's taste information from dialog \cite{Mera01}. The learning method consists of four following parts:(1)Directly expression according to the degree of like/dislike, (2)Decision Making based on the speaker's environment, (3)Relevance to the degree of displeasure to the object, and (4)Backward calculation from the emotional expression. The method seems to be effective in theory, but the procedure (2) is still an theoretical query into emotions, because it does not include the reasoning model. In this paper, the learning method employs Fuzzy Petri Net (FPN) to find an appropriate $FV$ for a word which has the unknown $FV$. The $FV$ given by the method can change the value itself according to the meaning or intent of discourse in a dialog. 

\begin{enumerate}
\item Directly Expression\\
For the extraction of $FV$ from the utterance, we should pay attention to the words including ``like'' and ``dislike''. These words are used to find one's impression about some objects. In the method, when the predicate in the sentence includes like/dislike, $FV$ for the object is set to a positive/negative value.
\item Change of Speaker's Feeling in his/her situation (Decision Making based on the speaker's environment)\\
$FV$ increases naturally if an object is something useful or is favorable to the agent. On the contrary, it decreases, if an object is something harmful or is unfavorable to the agent. $FV$ for a predicate of an event is assigned a pre-determined numerical value in $[0, 1]$. However, the method in \cite{Mera01} is unclear, because the 3rd rule, called `Favorable Value Changing Situation', does not have the reasoning model. 
\item Relevance to displeasure\\
An object that people feels in something displeasant tends to be disliked because it associates the past displeasant events. If a person encounters some unpleasant events, he/she will hate the objects in such an event. The $FV$ is reduced by such an idea, when the word appears in a unpleasant utterance.
\item Backward calculation\\
If the emotional expression includes in the sentence, $FV$ is given by the backward calculation of EGC.
\end{enumerate}

\subsection{Improvement of Learning Method}
\label{sec:ChangeFeeling}
The change of speaker's feeling originates from the exchange of ideas via conversation. Because EGCs are the equations to measure the emotion value in the form of case frame representation in a dialog, the reasoning rule of the equation is given in the form of FPN model as shown in Fig.\ref{fig:FPN_models}. However, the reasoning model has only 2 types, Type 1 and Type 2, because the EGC equation is a simple rule. In case of Type 2, the consequent parts in the rule are transformed to divide into 2 or more rules which are in a consequent part, and then we can consider the Type 1 rule only.
Table \ref{tab:FPNrules} shows the FPN rules for EGC in case frame representation. The $\mu$ in Table \ref{tab:FPNrules} is selected from the transition cost in MSTN in Fig.\ref{fig:MSTN_EGC} and the classification of generated emotion in Table\ref{tab:classofemotion}.

\begin{table}[tb]
\caption{FPN rules in case frame representation}
\label{tab:FPNrules}
\scalebox{1.0}[1.0]{ 
\begin{tabular}{p{0.3cm}p{0.1cm}l}
$R_{1}$  &:&IF S and V THEN ${\rm LIKE}, CF = \mu_{1}$\\
$R_{2}$  &:&IF S and V and OF THEN ${\rm LIKE}, CF = \mu_{2}$\\
$R_{3}$  &:&IF S and V and OT THEN ${\rm LIKE}, CF = \mu_{3}$\\
$R_{4}$  &:&IF S and V and OM THEN ${\rm LIKE}, CF = \mu_{4}$\\
$R_{5}$  &:&IF (S or OS) and V and OS THEN ${\rm LIKE}, CF = \mu_{5}$\\
$R_{6}$  &:&IF S and V and O THEN ${\rm LIKE}, CF = \mu_{6}$\\
$R_{7}$  &:&IF V and O and (OF or OT) THEN ${\rm LIKE}, CF = \mu_{7}$\\
$R_{8}$  &:&IF V and O and (OT or OF) THEN ${\rm LIKE}, CF = \mu_{8}$\\
$R_{9}$  &:&IF V and O and OM THEN ${\rm LIKE}, CF = \mu_{9}$\\
$R_{10}$ &:&IF V and O THEN ${\rm LIKE}, CF = \mu_{10}$\\
$R_{11}$ &:&IF O and OC THEN ${\rm LIKE}, CF = \mu_{11}$\\
\end{tabular}
}
\vspace{-0.3cm}
\end{table}

In Table \ref{tab:FPNrules}, $R_{5}$, $R_{7}$ and $R_{8}$ rules are divided into 2 rules, respectively. That is, $R_{51}$, $R_{52}$, $R_{71}$, $R_{72}$, $R_{81}$, and $R_{82}$ are given as follows.

{\small
\begin{tabular}{p{0.25cm}p{0.1cm}l}
$R_{51}$ &:&IF S and V THEN LIKE, $CF = \mu_{51}$\\
$R_{52}$ &:&IF OS and V THEN LIKE, $CF = \mu_{52}$\\
$R_{71}$ &:&IF V and O and OF THEN LIKE, $CF = \mu_{71}$\\
$R_{72}$ &:&IF V and O and OT THEN LIKE, $CF = \mu_{72}$\\
$R_{81}$ &:&IF V and O and OF THEN LIKE, $CF = \mu_{81}$\\
$R_{82}$ &:&IF V and O and OT THEN LIKE, $CF = \mu_{82}$\\ \\
\end{tabular}
}

The main role of learning method of $FV$ is the solution for the unknown $FV$, because the $FV$ is set to $0.5$ when the $FV$ is not given the degree of like/dislike to the corresponding word. On the other hand, even if the $FV$ has been already known, it is suitable to change its value according to the situation in emotional environment.

As mentioned in Section \ref{sec:FavoriteValue}, the range of $FV$ is $[-1.0, 1.0]$. Because the unknown $FV$ is $0.5$, we consider that the center of emotion value is $0.0$ and the positive/negative emotional space expands to 1.0 bidirectionally. Then, the normalized range and the positive/negative sign are required to backward calculation.

\subsubsection{$FV$ is unknown}
The token in $P_{ji}$ is the agreement value as the antecedent part in Type1 in Fig.\ref{fig:FPN_1} and then, the minimum value among them is calculated by Eq.(\ref{eq:Min}). If the place which takes the minimum value have an unknown $FV$, the agreement value is $0.5$. In the case, the $FV$ corresponding the word will be modified by Eq.(\ref{eq:DeltaFV1}).

\begin{equation}
\Delta FV_{min}=\frac{EV-y_{k}}{\mu_{i}}
\label{eq:DeltaFV1}
\end{equation}
where $P_{k}$ is the output of the rule, $P_{ji_{min}}$ is the place which takes the minimum value, and $EV$ is the user's real emotion value. The $FV$ update is implemented by Eq.(\ref{eq:FVmod1}). $\eta$ is the learning parameter.
\vspace{-0.3cm}
\begin{equation}
FV=FV+\eta \Delta FV_{min}
\label{eq:FVmod1}
\end{equation}

If the $FV$ of place $P_{ji_{min}}$ is known, but the other places have unknown $FV$, the $FV$ is modified by Eq.(\ref{eq:DeltaFV2}).
\begin{equation}
\Delta FV_{\mu}=\frac{EV-y_{u}\cdot\mu_{i}}{\mu_{i}} 
\label{eq:DeltaFV2}
\end{equation}
where $u (\neq i_{min})$ is the place with unknown $FV$. The $FV$ update is implemented by Eq.(\ref{eq:FVmod2}).

\begin{equation}
FV=FV+\eta \Delta FV_{u}
\label{eq:FVmod2}
\vspace{-0.3cm}
\end{equation}

\subsubsection{$FV$ is already known}
Even if the unknown $FV$ does not included in the sentence, sometimes the learning of $FV$ is required, and then the $\mu$ will be changed according to current mood in the speaker's environment. In the case, the $FV$ corresponding the word will be modified by Eq.(\ref{eq:DeltaFV3}). That it, Eq.(\ref{eq:DeltaFV3}) change only $FV$ which takes the minimum value and does not modify the other $P$. The modification rule is executed except the case of 2 words of negative $FV$s.
\begin{eqnarray}
\Delta FV_{min}=\left\{
\begin{array}{ll}
\frac{EV-y_{k}}{\mu_{i}} & if\;P = P_{ji_{min}}\\
0 & otherwise\\
\end{array}
\right.
\label{eq:DeltaFV3}
\end{eqnarray}

\section{Conclusion}
\label{sec:ConclusiveDiscussion}
The learning method of speaker's taste information from dialog in \cite{Mera01} was divided into four types and was implemented by using grammatical knowledge and already known favorable values. The three types except ``Change of Speaker's Feeling in his/her situation'' can find the appropriate $FV$, even if $FV$ of the word is unknown. However, it is difficult for the learning method in \cite{Mera01} to perceive the delicate change of mode in conversation, because EGC can measure the emotion value by $FV$ of word in a dialog. MSTN can measure not only the temporary change of emotion but also long-term variation. The learning method proposed in this paper employs Fuzzy Petri Net (FPN) to find an appropriate $FV$ to a word which has the unknown $FV$. The empirical studies are required to be fine tuning for the FPN rules and the definition of $\mu$.

\section*{Acknowledgment}
This work was supported by JSPS KAKENHI Grant Number 25330366.

\end{document}